%% file: arxiva.tex
\documentclass{article}

\usepackage[preprint]{neurips_2026}

\usepackage{wrapfig}
\usepackage{tikz}
\usetikzlibrary{positioning, fit, calc, shapes.geometric}

\usepackage[utf8]{inputenc} 
\usepackage[T1]{fontenc}    
\usepackage{hyperref}       
\usepackage{url}            
\usepackage{booktabs}       
\usepackage{amsfonts}       
\usepackage{nicefrac}       
\usepackage{microtype}      
\usepackage{xcolor}         
\usepackage{graphicx}
\usepackage{amsmath}

\definecolor{updatecolor}{RGB}{255, 102, 102}

\def\our{GLADOS}

\title{Mind the Gap: Geometrically Accurate Generative Reconstruction from Disjoint Views}

\author{%
  {\bf Grzegorz Wilczyński} \\
  Jagiellonian University \\
  IDEAS Research Institute\\
  \texttt{grzegorz.wilczynski@doctoral.uj.edu.pl} \\
  \and
  {\bf Miko{\l}aj Zielinski} \\
  Poznań University of Technology \\
  \and
  {\bf Bartosz \'Swirta} \\
  Warsaw University of Technology \\
  IDEAS Research Institute\\
  \and
   {\bf Dominik Belter} \\
  Poznań University of Technology \\
  \and
   {\bf Przemys{\l}aw Spurek} \\
  Jagiellonian University \\
  IDEAS Research Institute\\
}

\begin{document}

\maketitle

\begin{figure}[h!]
  \centering
  \vspace{-1cm}
  \includegraphics[width=\linewidth]{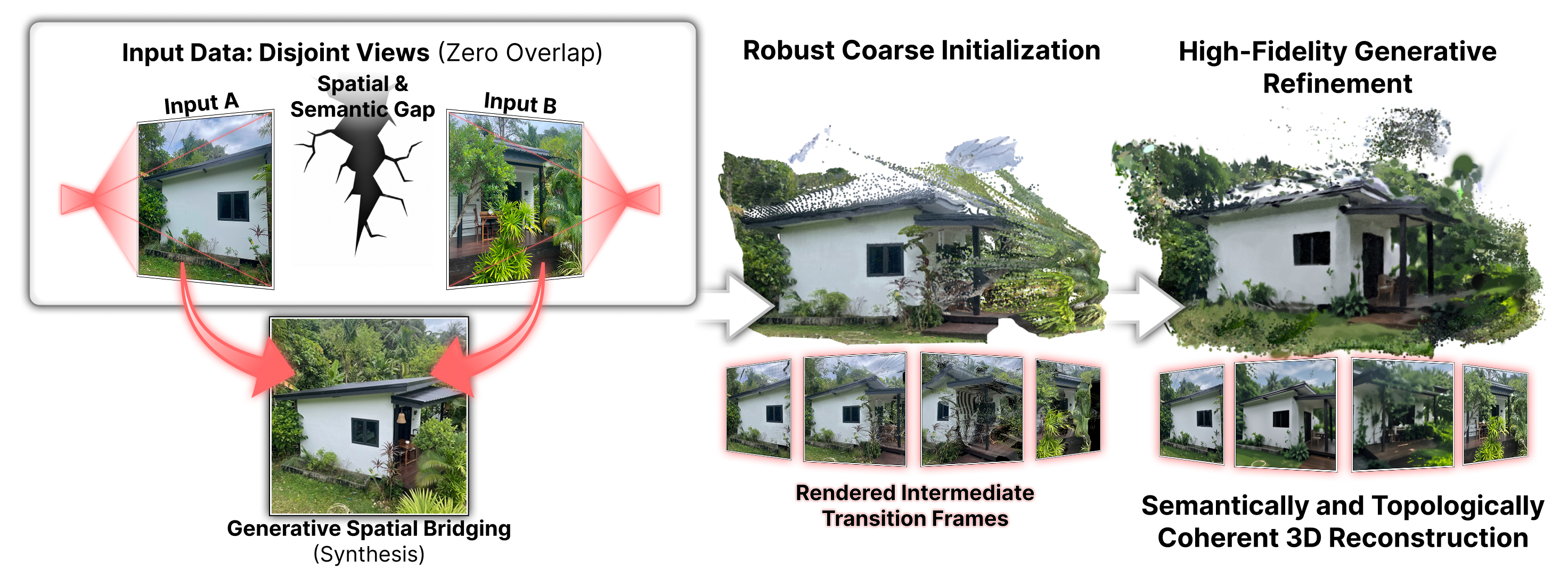}
    \vspace{-0.5cm}
  \label{fig:teaser}
  \caption{\textbf{\our{} Framework for Generative Reconstruction from Disjoint Views.} Our pipeline solves 3D reconstruction across zero-overlap ``spatial \& semantic gaps.'' First, \textit{Generative Spatial Bridging} uses VLM reasoning to synthesize intermediate transition frames, establishing a continuous visual link. Next, \textit{Robust Coarse Initialization} builds an initial 3D scaffold from this augmented sequence. Finally, \textit{High-Fidelity Generative Reconstruction} refines the geometry into a topologically coherent 3D model, enforcing original observations as hard constraints while hallucinating plausible unobserved regions.}
  \vspace{-0.5cm}
\end{figure}

\begin{abstract}
    3D vision systems are fundamentally constrained by their reliance on visual overlap: reconstruction methods require it for geometric alignment, while generative models use it to enforce multi-view consistency. This limitation is particularly acute in real-world scenarios such as distributed swarm robotics or crowd-sourced data collection, where capturing overlapping perspectives, both in terms of spatial and appearance overlap, is often impossible.
    We introduce \textbf{Generative Reconstruction from Disjoint Views} as a new paradigm, establish a comprehensive dataset, and propose specialized evaluation metrics for zero-overlap scenarios. Our benchmarking demonstrates that existing state-of-the-art methods fail catastrophically on this task, producing disconnected geometries or semantically incoherent reconstructions.
    To address these limitations, we propose \our{}, a general, modular framework that operates through three stages: \textit{(1) Generative Bridging}, where foundation models synthesize intermediate perspectives to connect disjoint inputs; \textit{(2) Robust Coarse 3D Reconstruction}, that establish coarse geometric scaffold via global alignment which absorbs local contradictions from generative process; and \textit{(3) Iterative Context Expansion and Consistency Optimization} to fill missing regions and unify the reconstruction. As an architecture-agnostic framework, \our{} enables seamless integration of future advances in generation, reconstruction, and inpainting.
    The source code is available at: \url{https://github.com/gwilczynski95/GLADOS}.

\end{abstract}

\input{sections/1-Introduction}
\input{sections/2-Motivation}
\input{sections/3-Problem}

\begin{figure}[t]
  \centering
  \includegraphics[width=\textwidth]{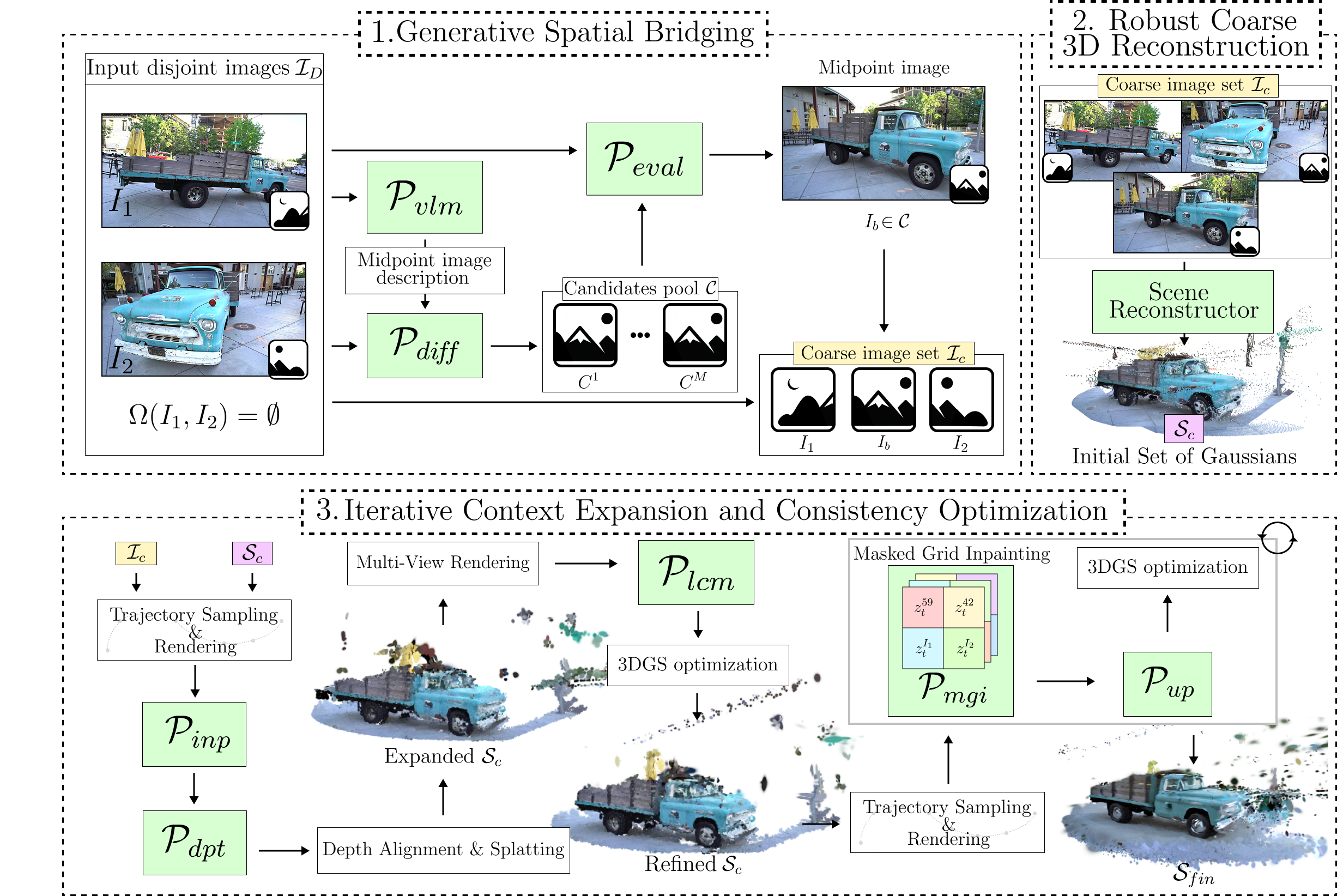}
  \vspace{-0.35cm}
  \caption{The pipeline consists of three main stages. First, \textit{Generative Spatial Bridging} synthesizes a semantic anchor image to connect disjoint input views, creating $\mathcal{I}_{c}$. Second, \textit{Robust Coarse 3D Reconstruction} aligns this augmented set to initialize a coarse 3D Gaussian Splatting scene ($\mathcal{S}_{c}$). Finally, iterative expansion and consistency optimization expand the scene using $\mathcal{P}_{inp}$ and $\mathcal{P}_{lcm}$, while our custom \textit{Anchored Grid Inpainting} ($\mathcal{P}_{mgi}$) and super-resolution ($\mathcal{P}_{up}$) iteratively eliminate residual holes to yield a metrically accurate, completed reconstruction ($\mathcal{S}_{fin}$).}
  \label{fig:method}
  \vspace{-0.5cm}
\end{figure}

\input{sections/4-Related_works}
\input{sections/5-Method}
\input{sections/6-Benchmark}
\input{sections/7-Results}
\input{sections/8-Conclusions}

\bibliographystyle{unsrt}

\newpage
\appendix

\section{Metrics}

\paragraph{GeCo (Geometric Consistency Metric)} is a differentiable, geometry-grounded metric designed to detect geometric deformation and occlusion-inconsistency artifacts in static scenes. It combines residual motion and depth priors to produce interpretable, dense consistency maps. Specifically; Motion Map highlighting motion inconsistencies between object and camera-induced motion, Structure Map defined as the residual between the predicted and reprojected depth and Fused Map that fuses two previous maps into single scale-invariant error map. By quantifying geometric consistency, GeCo enables systematic benchmarking of generative models, exposing common failures of 3D consistency like morphing geometries and texture flickering.

\paragraph{MEt3R (Measuring Multi-View Consistency in Generated Images)} is designed to evaluate the 3D consistency between pairs of generated multi-view images, independent of image quality, content, or camera poses. It leverages DUSt3R to create dense 3D reconstructions from image pairs, then warps and compares feature maps to compute a similarity score that is robust to view-dependent effects like lighting changes. Thus MEt3R provides a measure for structural and semantic consistency between subsequent images, crucial in our scenario, where reconstruction is created with weak ground truth anchoring. 

\section{Implementation Details}
\label{sec:implementation_details}

All experiments, including dataset generation, 3D reconstruction, and refinement optimization, were conducted on a single NVIDIA A100 GPU with 40GB of VRAM. Our \our{} framework orchestrates several state-of-the-art foundation models to achieve robust geometric lifting and generative refinement, some of which are accessible via 3rd-party API. The specific model instantiations for each variable in our pipeline are detailed below:

\paragraph{Generative Spatial Bridging.} 
For the synthesis of the semantic anchor, the initial Vision-Language Model ($\mathcal{P}_{vlm}$) that acts as the reasoning and prompting engine is implemented using Gemini 3.1 Flash \cite{gemini}. The multi-conditional diffusion model ($\mathcal{P}_{diff}$) tasked with generating the diverse pool of candidate images $\mathcal{C}$ is Nano Banana 2 \cite{nanobanana}. To evaluate and select the optimal anchor $I_{b}$, the zero-shot reward model ($\mathcal{P}_{eval}$) utilizes Qwen2-VL-7B-Instruct \cite{qwen}.

\paragraph{Robust Coarse 3D Reconstruction.} 
To extract the initial coarse geometric scaffold ($\mathcal{S}_c$) from the sparse triplet $\mathcal{I}_c$, the unconstrained dense pointmap regression model ($\mathcal{P}_{geo}$) is implemented using Dust3R \cite{wang2024dust3r}.

\paragraph{Iterative Context Expansion and Consistency Optimization.} 
During the warp-and-inpaint context expansion phase, the RGB inpainting prior ($\mathcal{P}_{inp}$) is powered by Fooocus \cite{fooocus}. For structural depth prediction and alignment, the monocular metric depth estimator ($\mathcal{P}_{dpt}$) utilizes DepthPro \cite{depthpro}. The Multiview Consistency Sampling (MCS) is driven by the Latent Consistency Model ($\mathcal{P}_{lcm}$) built upon the Stable Diffusion architecture \cite{lcm}.

\paragraph{Generative 3D Refinement via Anchored Grid Inpainting.} 
In the final iterative refinement loop, the grid-based diffusion prior ($\mathcal{P}_{mgi}$) used to inpaint the $2 \times 2$ anchored composite grids is based on Stable Diffusion v1.5 \cite{sd}. Finally, the super-resolution prior ($\mathcal{P}_{up}$) employed to restore high-frequency textural fidelity to the inpainted patches prior to 3DGS integration is Real-ESRGAN \cite{realesrgan}.

\subsection{Hyperparameters}
\label{subsec:hyperparameters}

\paragraph{Generative Spatial Bridging.} 
During the creation of the semantic anchor, we sample $M=3$ diverse candidate midpoint images from the multi-conditional diffusion model $\mathcal{P}_{diff}$. The VLM evaluator $\mathcal{P}_{eval}$ analyzes these three candidates, and the highest-scoring image is deterministically selected as the optimal anchor $I_{b}$ to initialize the coarse coordinate space.

\paragraph{Iterative Context Expansion and Consistency Optimization.}
For the intermediate scene completion, we adopt hyperparameters aligned with \cite{wang2025vistadream}. We establish an interpolated camera trajectory for the iterative warp-and-inpaint procedure. Each time new Gaussian primitives are unprojected into the scene from $\mathcal{P}_{inp}$, the scene undergoes a brief 256-step optimization. For the Multiview Consistency Sampling (MCS), we uniformly render $N = 8$ views from the coarse scaffold at a resolution of $512 \times 512$. We add $T = 10$ steps of noise using a standard 50-step diffusion schedule. The Latent Consistency Model ($\mathcal{P}_{lcm}$) predicts the noise without Classifier-Free Guidance. To strictly enforce multi-view consistency without collapsing the geometry, the intermediate Gaussian field is optimized for 2560 steps utilizing a reduced spatial learning rate for the Gaussian centers $10^{-4}$ \cite{wang2025vistadream}.

\paragraph{Generative 3D Refinement via Anchored Grid Inpainting.}
Our final, topology-grounding refinement loop is executed for a total of 5 iterative cycles. During each cycle, the grid-based diffusion prior $\mathcal{P}_{mgi}$ processes the $2 \times 2$ anchored composite images using a partial denoising schedule that finishes at $0\%$ noise. To ensure convergence, the starting noise level is linearly annealed across the 5 cycles: it begins at $20\%$ for the first cycle and drops down to $0.05\%$ by the final cycle. Once the inpainted patches are enhanced by $\mathcal{P}_{up}$ and injected into the masked regions of the scene, the updated 3D Gaussian field undergoes 300 optimization iterations. Throughout this entire refinement stage, standard default 3D Gaussian Splatting optimization hyperparameters are maintained.

\section{Qualitative Ablation Analysis}
\label{sec:qualitative_ablation}

\begin{figure}[h]
  \centering
  \includegraphics[width=\linewidth]{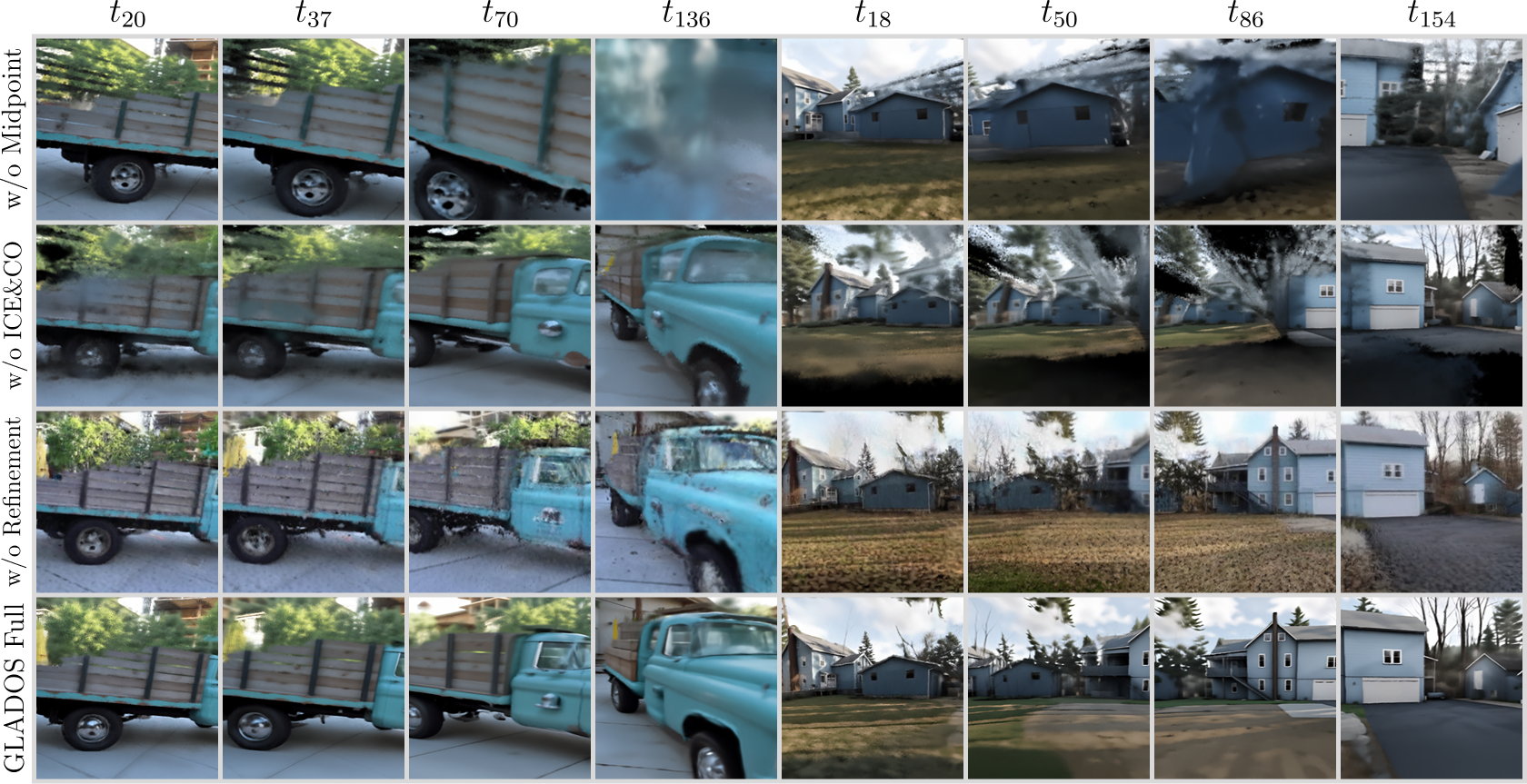}
  \vspace{-0.35cm}
  \caption{\textbf{Qualitative Ablation Results}. The left four columns show views of the Truck scene from the Tanks and Temples dataset, while the right four columns depict results for Scene 185 from the RealEstate10k dataset.}
  \label{apdx:ablation}
\end{figure}

To complement the quantitative results presented in the main text in Figure \ref{fig:figure_5}, Figure \ref{apdx:ablation} provides a visual comparison of our ablation configurations on the Truck scene (left) and Scene 185 from RE10K (right). 

Without generative spatial bridging, presented as \textit{w/o Midpoint}, the system suffers complete registration failure on the Truck scene, resulting in missing geometry and yields fragmented, poorly aligned structures in Scene 185. Bypassing the iterative expansion phase (\textit{w/o ICE\&CO}) might lead to geometric collapse on scenes with significant translation between input views.

Halting the pipeline before our anchored grid inpainting, presented as \textit{w/o Refinement}, yields individual renders of high visual quality, which explains its FID score, but this 2D appearance masks 3D structural defects. Visually, the underlying geometry suffers from residual unobserved holes, missing structural patches, and a lack of precise boundary definition. In contrast, the full \our{} pipeline successfully leverages all modules to resolve these geometric artifacts, delivering coherent 3D reconstructions.

\section{Additional results}

We present additional examples that extend the results presented in Figure ~\ref{fig:figure_5} and provide corresponding video files in supplementary materials.

\begin{figure}[h]
  \centering
  \includegraphics[width=\linewidth]{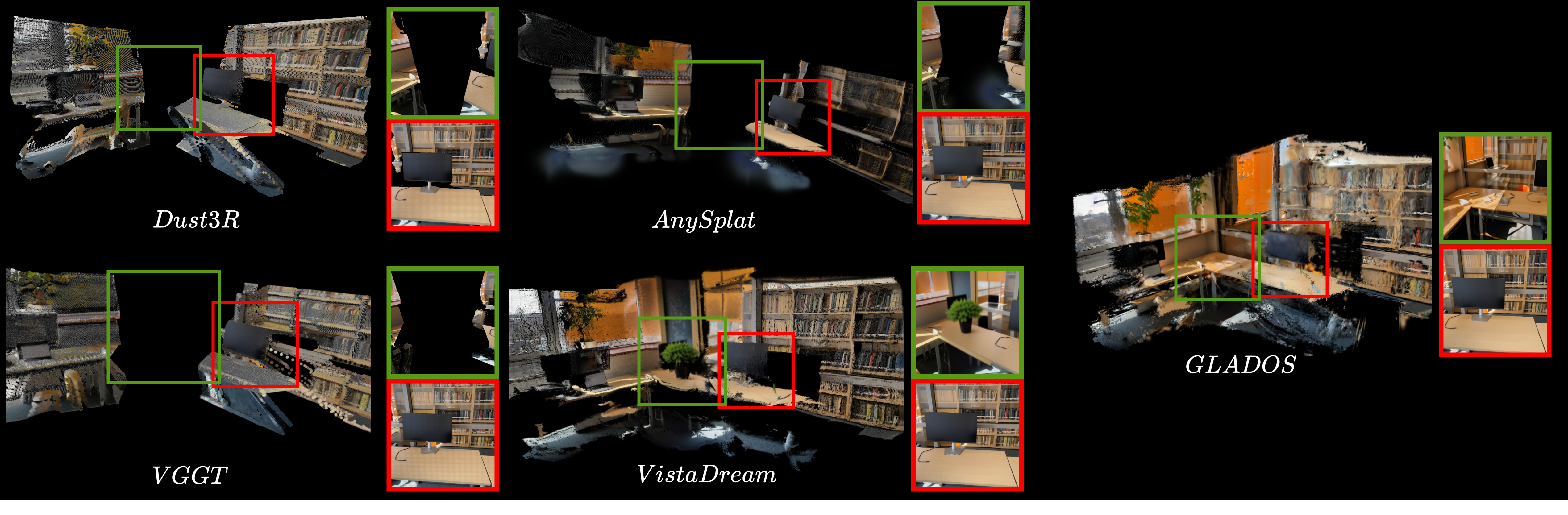}
  \vspace{-0.35cm}
  \caption{\textbf{Comparison of the final reconstruction between \our{} and other methods.} Purely reconstruction methods (\textit{Dust3R}, \textit{VGGT}, \textit{AnySplat}) produce solid results for seen regions (red), they cannot fill in the gaps that are the result of disjoint constraint (green). \textit{VistaDream} can generate missing geometry, but despite being visually good it is semantically improbable, half of the table behind glass wall and very small monitor behind the plant. Our method produces visually coherent results that are probable in the context of the given scene.}
  \label{apdx:fig:quality}
\end{figure}

\begin{figure}[h]
  \centering
  \includegraphics[width=\linewidth]{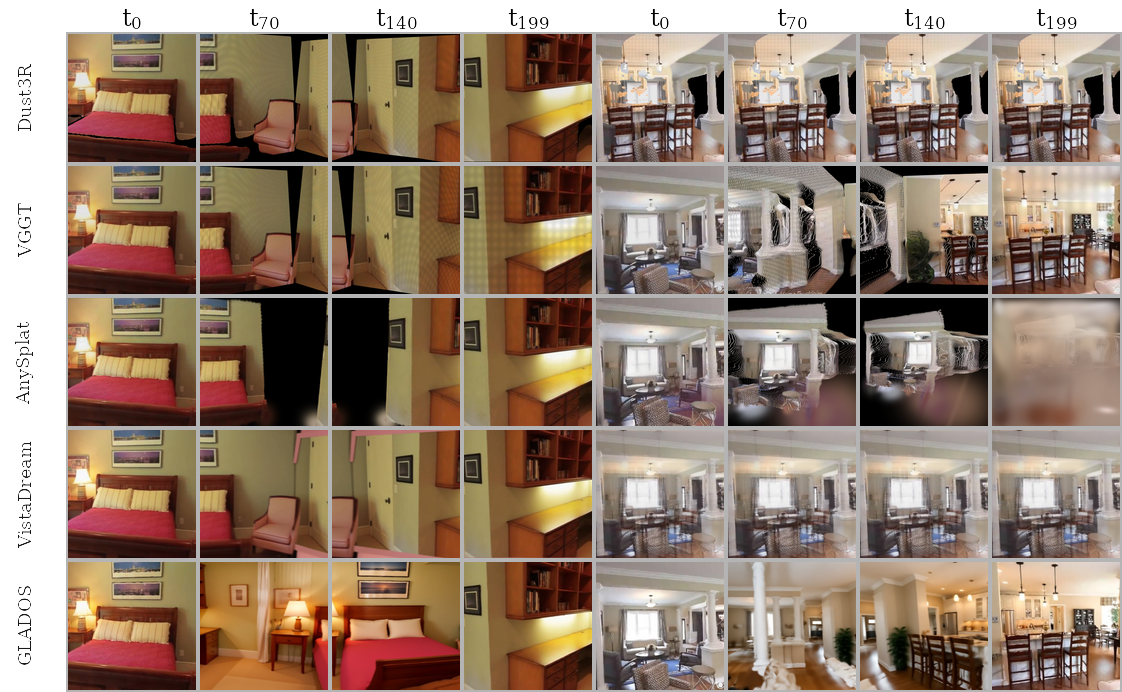}
  \vspace{-0.35cm}
  \caption{On the left scene 15 from \textit{RE10K}, on the right scene 17 from \textit{RE10K}}
  \label{apdx:consistency_2}
\end{figure}

\begin{figure}[h]
  \centering
  \includegraphics[width=\linewidth]{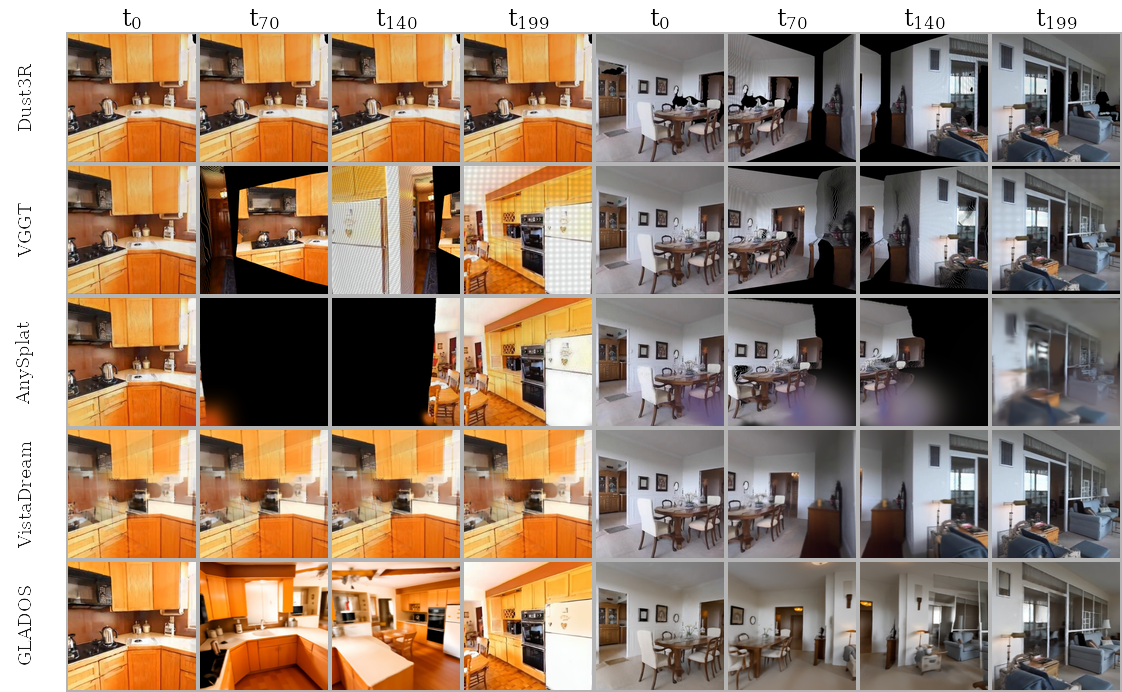}
  \vspace{-0.35cm}
  \caption{On the left scene 25 from \textit{RE10K}, on the right scene 357 from \textit{RE10K}}
  \label{apdx:consistency_3}
\end{figure}

\textit{Dust3R}, \textit{VGGT} and \textit{AnySplat} struggle with reconstruction of unseen regions, with \textit{Dust3R} failing on few scenes, as seen in Figure \ref{apdx:consistency_2} and Figure \ref{apdx:consistency_3} where it returns geometry estimation from single camera or blends two views into one. \textit{VGGT} exhibits noticeable visual artifacts (grid-like textures), whereas \textit{AnySplat} renders the most empty spaces and fails to properly reconstruct second view. \textit{VistaDream} can generate missing geometry to some extend lacking overall robustness. Our method, \our{} produces visually coherent results that are probable in the context of the given scene.

\begin{figure}[h]
  \centering
  \includegraphics[width=\linewidth]{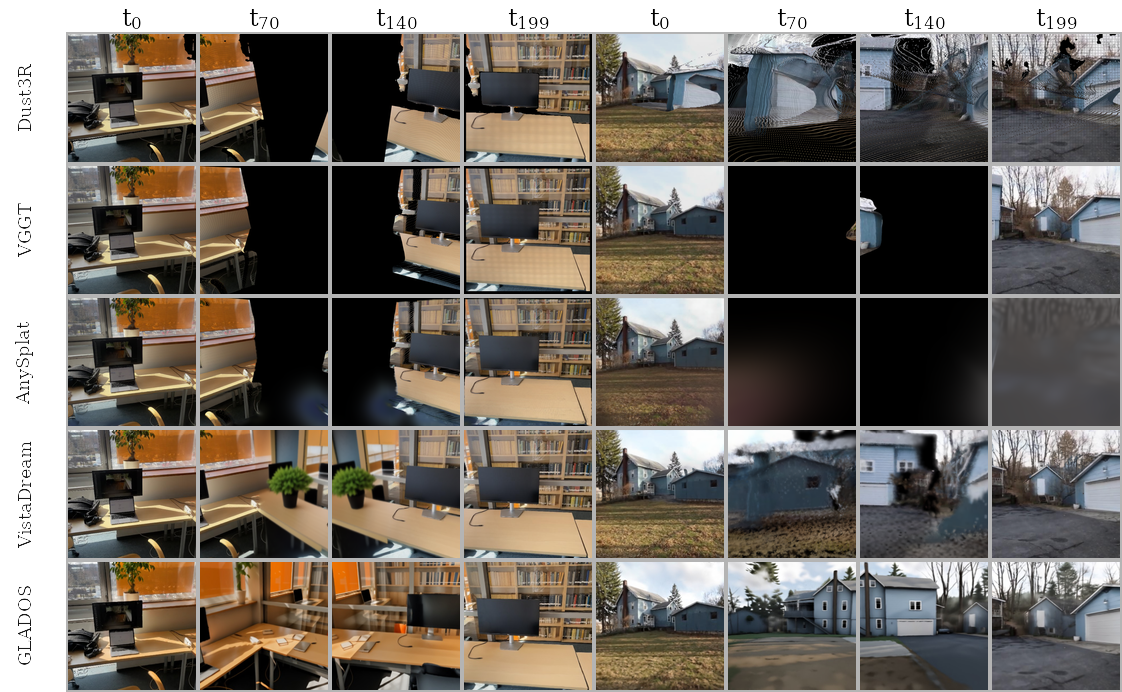}
  \vspace{-0.35cm}
  \caption{On the left self captured scene \textit{library}, on the right scene 185 from \textit{RE10K}}
  \label{apdx:consistency}
\end{figure}

\newpage

\section{Broader Impact}

This work introduces a new paradigm for coherent 3D reconstruction from spatially disjoint observations, enabling reconstruction in scenarios where traditional overlap-based methods fail. The proposed framework has potential positive impact across several domains. In robotics and autonomous exploration, the ability to infer coherent spatial structure from sparse and non-overlapping observations may improve mapping efficiency in single and multi-agent exploration systems. In consumer applications, the method could facilitate low-effort generation of virtual environments and digital twins from casually captured imagery. The framework may also benefit large-scale cultural heritage preservation and crowd-sourced reconstruction, where complete visual coverage is often unavailable.

At the same time, the ability to create realistic spatial reconstructions from sparse observations could potentially be misused to generate deceptive or misleading virtual environments. To mitigate these risks, our work explicitly distinguishes between observed and generated content and emphasizes that the generated intermediate regions should be treated as probabilistic scene completions rather than ground-truth geometry. We additionally provide evaluation metrics designed to measure geometric consistency and reconstruction reliability in the proposed zero-overlap setting.


\end{document}

%% file: sections/1-Introduction.tex
\section{Introduction}
The landscape of 3D computer vision has been fundamentally transformed by the advent of Neural Radiance Fields (NeRFs) \cite{mildenhall2020nerf} and the subsequent leap in rendering efficiency introduced by 3D Gaussian Splatting \cite{kerbl20233d}. Building upon these core representations, recent foundation models such as Dust3r \cite{wang2024dust3r} have shifted the paradigm toward unconstrained, feed-forward geometry estimation. These dense architectures, alongside instant 3D generation frameworks such as AnySplat \cite{jiang2025anysplat} and Splatt3r \cite{smart2024splatt3r}, can rapidly reconstruct high-fidelity scenes without relying on classical camera calibration or strict structure-from-motion algorithms.

However, the success of both traditional and neural reconstruction pipelines depends on a critical assumption: a non-empty visual overlap between the input images. To compute feature correspondences, align coordinate systems, and optimize scene parameters, these models fundamentally require shared visual information. In practical applications such as swarm robotics \cite{cieslewski2018data}, autonomous surveillance \cite{loy2010time}, or unconstrained crowd-sourcing \cite{heinly2015reconstructing}, cameras often capture disjoint views to maximize spatial coverage. When presented with zero-overlap image pairs, state-of-the-art geometry estimators fail to register coordinate frames. This results in disconnected and unaligned geometric clusters, a structural challenge that even advanced global alignment strategies like Align3r \cite{lu2025align3r} or efficient solvers like Fast3R \cite{yang2025fast3r}, and Easi3r \cite{chen2025easi3r} cannot easily resolve without shared pixels.

Conversely, the recent surge in 3D generative models offers an alternative approach focused on creation rather than strict extraction. Frameworks such as VistaDream \cite{wang2025vistadream}, LucidDreamer \cite{chung2025luciddreamer}, and DreamScape \cite{zhao2025dreamscape} demonstrate a remarkable ability to hallucinate entire 3D scenes from minimal prompts. Furthermore, novel view synthesis and volume generation methods such as ViewCrafter \cite{yu2025viewcrafter} and WVD \cite{zhang2025world} can synthesize plausible unobserved spaces. While these generative priors excel at producing diverse content, they struggle to enforce strict geometric constraints across multiple disjoint observations simultaneously. They typically prioritize aesthetic coherence over the structural fidelity required to reconstruct the original inputs perfectly, often leading to geometric distortions or topological inconsistencies when forced to merge independent views.

To address this fundamental limitation, we define a novel computer vision task:  {\bf Generative Reconstruction from Disjoint Views}. 
This task requires a system to synthesize a single, unified 3D environment from entirely non-overlapping images. It must treat the original disjoint observations as hard geometric constraints while seamlessly generate the intermediate space using learned spatial priors. To solve this challenge, we introduce \our{}, a framework that integrates the rigor of geometric reconstruction with the flexibility of generative models. \our{} utilizes {\em Generative Spatial Bridging} via foundation models to establish a logical anchor image that visual connects the disjoint inputs. Subsequently, we employ a 3D-consistent masking and filling strategy called {\em Iterative Context Expansion and Consistency Optimization} to progressively expand the visual context. This provides the necessary synthetic overlap to ground a highly accurate 3D Gaussian Splatting optimization.
The main contributions of our work are summarized as follows:
\vspace{-0.2cm}
\begin{itemize}
    \item We formalize the novel task of Generative Reconstruction from Disjoint Views and introduce a specialized benchmark with evaluation metrics tailored to assess both geometric accuracy and generative consistency in zero-overlap scenarios.
\vspace{-0.15cm}
    \item We propose \our{}, a pioneering three-stage framework that effectively leverages {\em Generative Spatial Bridging} and {\em Iterative Context Expansion and Consistency Optimization} to synthesize unobserved spatial connections between disconnected image pairs.
\vspace{-0.15cm}
    \item We demonstrate, through extensive benchmarking, that GLADOS outperforms state-of-the-art reconstruction and generative baselines, successfully unifying the reconstruction while preserving the structural fidelity of the original views.
\vspace{-0.2cm}
\end{itemize}

%% file: sections/2-Motivation.tex
\section{Motivation}
\vspace{-0.2cm}
The strict requirement for visual overlap in traditional 3D reconstruction pipelines creates a severe bottleneck for real-world deployment. While existing algorithms thrive in controlled environments where a single user slowly orbits an object, they are inherently unsuited for dynamic, distributed, or casual data capture. We identify three practical scenarios in which reconstruction from disjoint views naturally arises.

\textbf{Swarm Robotics and Autonomous Exploration.} In time-critical scenarios such as search-and-rescue or disaster response, deploying a swarm of drones is well-established as the most efficient way to map a large area \cite{erdelj2017wireless,recchiuto2018post}. To maximize spatial coverage and battery efficiency, individual agents naturally follow divergent trajectories, capturing multi-directional views. Consequently, the swarm produces disjoint visual data with massive spatial gaps. While state-of-the-art multi-robot SLAM and structure-from-motion pipelines \cite{tian2022kimera,lajoie2020door,schmuck2019ccm,evangelidis2021revisiting} excel at map merging and geometry initialization, they strictly depend on loop closures, shared visual landmarks, and overlapping perspectives. A robust 3D system must be able to infer the topological relationships among these isolated agents to construct a unified command map, thereby eliminating the burden of forcing them to waste critical mission time on capturing overlapping terrain.

\textbf{Consumer-Level Capture and Virtual Tours.} The creation of virtual real estate tours or digital twins often relies on casual users capturing images with their smartphones in highly unconstrained environments \cite{chang2017matterport3d}. A typical user might photograph the front of a living room and then move on to the kitchen, completely omitting the connecting hallway or other transitional spaces. Traditional photogrammetry and structure-from-motion pipelines \cite{schonberger2016structure}, as well as modern few-shot neural rendering methods \cite{yang2023freenerf}, fail here due to a lack of dense visual overlap, treating the rooms as disconnected floating models. Generative reconstruction bridges this gap. Drawing inspiration from recent advances in multi-view 3D generation and generative scene completion \cite{hollein2023text2room, shi2023mvdream}, our approach enables the system to synthesize the missing architectural topology, providing a seamless 3D walkthrough experience based solely on sparse, disconnected anchor images.

\textbf{Unconstrained Crowdsourced Data.} When reconstructing landmarks or city scales from internet photo collections \cite{agarwal2011building, martin2021nerf}, the view distribution is heavily skewed toward popular, aesthetically pleasing angles. Tourists frequently photograph the ornate facade of a building and the courtyard behind it, but rarely capture the mundane side alleys required to physically connect them. This inevitably leads to fragmented 3D clusters and disconnected components in large-scale pipelines \cite{heinly2015reconstructing}. Contrary to other approaches which produce isolated geometric islands, our generative reconstruction paradigm synthesizes the unobserved intermediate regions, enabling continuous 3D models from unstructured crowd-sourced data.


%% file: sections/3-Problem.tex
\vspace{-0.2cm}
\section{Problem Formulation}
\label{sec:problem}
\vspace{-0.2cm}

To rigorously define the novel task of Generative Reconstruction from Disjoint Views, we first formulate the standard 3D reconstruction problem and highlight its fundamental limitations.

\paragraph{Classical 3D Reconstruction.} 
Given a set of input images $\mathcal{I} = \{I_1, I_2, \dots, I_N\}$ capturing a 3D scene, standard reconstruction aims to estimate a 3D representation $\mathcal{S}$ (e.g., a mesh, Neural Radiance Field, or 3D Gaussian Splatting) and the corresponding camera poses $\Pi = \{\pi_1, \pi_2, \dots, \pi_N\}$. Let $\mathcal{R}(\mathcal{S}, \pi_i)$ denote the differentiable rendering operator that projects the 3D scene $\mathcal{S}$ onto the 2D image plane at pose $\pi_i$. The objective is to minimize the photometric reconstruction error:
\begin{equation}
\mathcal{L}_{recon} = \sum_{i=1}^{N} \mathcal{D}(\mathcal{R}(\mathcal{S}, \pi_i), I_i), 
\end{equation}
where $\mathcal{D}$ is a distance metric, typically MSE or LPIPS. However, solving this optimization fundamentally requires visual overlap. Let $\Omega(I_i, I_j)$ represent the set of corresponding 3D points visible in both $I_i$ and $I_j$. Classical alignment and structure-from-motion strictly demand that for at least some pairs, $\Omega(I_i, I_j) \neq \emptyset$.

\textbf{The Disjoint View Constraint.}
In our unconstrained setup, we consider an extreme scenario where the input consists of a set of disjoint views, $\mathcal{I}_D = \{I_1, I_2, ... I_k\}$, capturing distinct regions of a scene. The primary constraint defining this problem is the zero-overlap condition, $\Omega(I_i, I_j) = \emptyset$, where $i,j \in K$, $i \neq j$. Under this condition, no geometric correspondences can be extracted. Consequently, traditional pose estimation fails to establish a relative transformation $T_{i \to j}$, and classical objective functions diverge, producing independent, unaligned local coordinate systems.

\textbf{Generative Reconstruction Task.}
We define Generative Reconstruction as the joint task of satisfying hard geometric constraints for observed regions while generating a structurally and semantically coherent space for the unobserved intermediate regions. Formally, we seek to find most probable latent geometry given sparse set of observations. Let $S$ be the latent geometry representation (e.g. neural radiance field, a voxel grid, or 3D Gaussians primitives), $\mathcal{D} = \{(\pi_{1}, I_1), \dots, (\pi_{K}, I_K)\}$ be the sparse set of $K$ known camera poses and ground-truth images, $\mathcal{V} = \{\pi_{K+1}, \dots, \pi_{N}\}$ be the dense set of novel cameras in unseen regions. The posterior probability of the geometry is expressed as $p(S | \mathcal{D}) \propto p(\mathcal{D} | S) p(S)$. Thus, minimization objective can be expressed as $-\log p(\mathcal{D} | \mathcal{S}) - \log p(\mathcal{S}).$ This can be split into two distinct parts (I) reconstruction term, (II) generative term. 

{\em Reconstruction term} enforces that obtained geometry matches provided sparse views under their camera poses. This is can be approximated via standard photometric loss $\mathcal{L}_{total} = \sum_{i \in K} \mathcal{L}_{recon}(\mathcal{S}, \pi_k, I_k)$. Because evaluating probability of given geometry is non trivial we approximate it with the {\em generative prior term} $\mathcal{L}_{gen}$, parameterized by a large multi-modal foundation model $\mathcal{P}_\theta$. This prior guides the synthesis of the unobserved volumetric space $\mathcal{S}_{unobs} = \mathcal{S} \setminus (\mathcal{S}_k: k \in {K})$, ensuring that the generated geometry smoothly bridges the disjoint observations in a plausible manner. 
The derived objective can be expressed as:
\begin{equation}
\mathcal{L}_{total} = \sum_{i \in K} \big( \mathcal{L}_{recon}(\mathcal{S}, \pi_k, I_k) \big) + \lambda \mathcal{L}_{gen}(\mathcal{S} \mid \mathcal{P}_\theta), \mbox{where $\lambda$ serves as balancing weight.}
\end{equation}


%% file: sections/4-Related_works.tex
\vspace{-0.3cm}
\section{Related Work}

Our research is positioned at the intersection of three rapidly evolving domains in computer vision: neural 3D reconstruction, foundation models for geometric alignment, and generative 3D synthesis. While traditional and feed-forward reconstruction methods excel at capturing high-fidelity geometry, they fundamentally rely on overlapping observations. Conversely, generative scene completion models demonstrate remarkable capabilities in generating unobserved spaces but often struggle to enforce rigid structural constraints across multiple independent anchors. In this section, we review recent advancements within these paradigms and highlight the critical limitations that our novel Generative Reconstruction framework directly addresses.

\textbf{Neural 3D Representations and Instant Reconstruction.}
The field of 3D computer vision has experienced a paradigm shift with the introduction of Neural Radiance Fields (NeRFs) \cite{mildenhall2020nerf}, which utilize coordinate-based multi-layer perceptrons for continuous volumetric scene representation. Subsequently, 3D Gaussian Splatting \cite{kerbl20233d} dramatically accelerated rendering speeds and enabled explicit geometric editing by modeling scenes as unstructured point clouds of anisotropic Gaussians. Building upon these efficient representations, recent efforts have focused on instant, feed-forward 3D reconstruction. Architectures such as AnySplat \cite{jiang2025anysplat}, Splatt3r \cite{smart2024splatt3r}, and TranSplat \cite{zhang2025transplat} have demonstrated the ability to predict dense 3D Gaussian primitives directly from sparse image sets in a single forward pass, bypassing the lengthy optimization typically required by traditional structure-from-motion. However, these methods intrinsically rely on shared visual context; when presented with disjoint views lacking overlap, they are unable to construct a unified global coordinate system.

\begin{wrapfigure}{r}{0.5\textwidth}
  \centering
  \vspace{-0.7cm}
  \includegraphics[width=\linewidth]{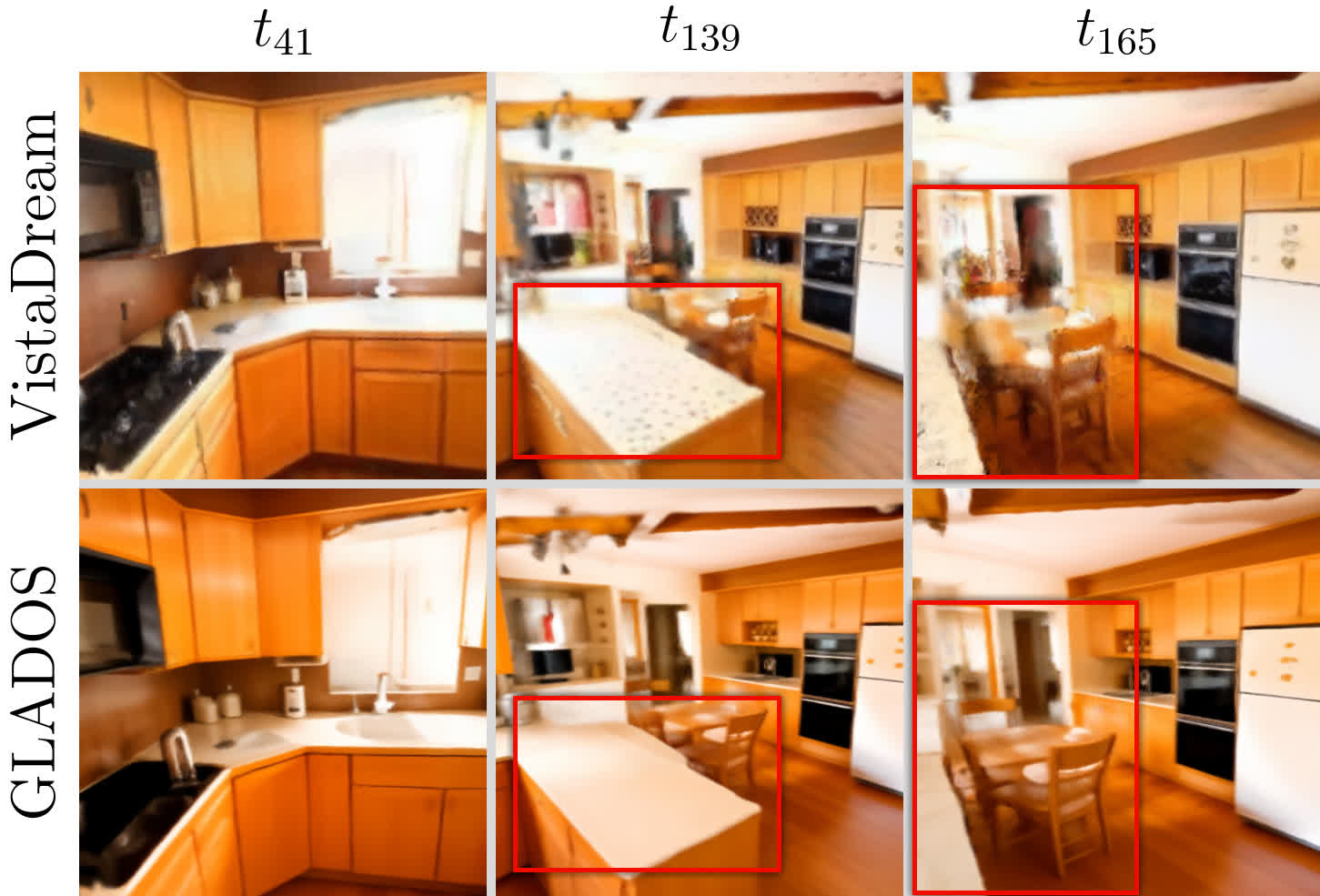} 
  \caption{Using the same input and midpoint images, VistaDream exhibits residual geometric voids ($t=139$) and textural blur ($t=165$). \our{}  eliminates these artifacts, producing a high-fidelity, complete 3D scene.}
  \vspace{-0.3cm}
  \label{fig:holes}
\end{wrapfigure}

\textbf{Foundation Models for Geometry and Alignment.}
Significant progress has also been made in learning robust, pixel-level geometric priors from massive datasets. Models like Dust3r \cite{wang2024dust3r} reframe 3D reconstruction as a dense point-matching and regression task, establishing strong local correspondences without explicit camera calibration. To tackle global scene alignment, sophisticated solvers and alignment frameworks such as Fast3R \cite{yang2025fast3r}, Easi3r \cite{chen2025easi3r}, Align3r \cite{lu2025align3r}, and VGGT \cite{wang2025vggt} have been proposed to efficiently register and merge multiple partial 3D reconstructions. In parallel, advances in spatial understanding, such as 3D-SAM \cite{kirillov2023segment}, have provided powerful tools for segmenting and structuring these extracted environments. Despite their robustness to noise and wide baselines, all of these state-of-the-art alignment strategies operate under the fundamental assumption of overlapping geometry. They fail to bridge the spatial gap when input sets share zero corresponding features.

\textbf{Generative 3D Synthesis and Scene Completion.}
To overcome the rigid constraints of traditional reconstruction, the community has increasingly turned to large-scale generative priors. Methods focusing on novel view synthesis and volume generation, such as ViewCrafter \cite{yu2025viewcrafter} and WVD (World Volume Diffusion) \cite{zhang2025world} leverage diffusion models to generate plausible observations beyond the original camera frustum. For large-scale environments, frameworks like VistaDream \cite{wang2025vistadream}, LucidDreamer \cite{chung2025luciddreamer}, DreamScape \cite{zhao2025dreamscape}, can synthesize entire, semantically rich 3D worlds from minimal text or image prompts. More closely related to our objective are 3D-consistent inpainting and outpainting techniques, such as Nerfiller \cite{weber2024nerfiller} and SceneTok \cite{asim2026scenetok}, which attempt to fill occlusions or missing data within a reconstructed scene. While highly creative, these purely generative approaches lack the structural rigor required to precisely anchor multiple independent, real-world observations. Our work directly addresses this gap by combining the strict geometric fidelity of neural reconstruction with the unconstrained synthesis capabilities of foundation models.


%% file: sections/5-Method.tex
\vspace{-0.2cm}
\section{\our{} Framework}
\label{sec:method}
\vspace{-0.2cm}


To tackle the Generative Reconstruction from Disjoint Views challenge introduced in Sec.~\ref{sec:problem}, we propose \our{}, a unified and modular framework that integrates the generative capabilities of foundation models with the geometric rigor of neural 3D reconstruction. The overall architecture of the pipeline is shown in Fig.~\ref{fig:method}. Although the formulation considers a general set of \(N\) input images, we focus here on the fundamental case of two disjoint views, \(\mathcal{I}_D = \{I_1, I_2\}\), where \(\Omega(I_1, I_2) = \emptyset\). Our framework operationalizes the generative prior \(\mathcal{P}_\theta\) as a cascading pipeline for synthesizing the unobserved space \(\mathcal{S}_{u}\). To ensure robustness and modularity, the pipeline is divided into three stages: {\em Generative Spatial Bridging}, {\em Robust Coarse 3D Reconstruction}, and {\em Iterative Context Expansion and Consistency Optimization}.

\begin{table}[t]
\caption{\textbf{Quantitative comparison} on the proposed disjoint-view reconstruction benchmark using only \textbf{two non-overlapping input images}.}
\label{tab:two_views}
\centering
\small
\setlength{\tabcolsep}{5pt}
\begin{tabular}{lcccccccc}
\toprule
Method & CLIP $\uparrow$ & FID $\downarrow$ & MET3R $\downarrow$ & $\text{GeCo}_M \downarrow$ & $\text{GeCo}_D \downarrow$ & $\text{GeCo}_F \downarrow$ & Photo $\downarrow$ & Rec. Fail $\downarrow$ \\
\midrule
AnySplat & 23.032 & 304.002 & 0.039 & 0.023 & 0.503 & 0.313 &\bf 0.077 & \bf 0 \\
VGGT & 22.751 & 275.245 & 0.043 & 0.029 & 0.109 & 0.075 & 0.132 & \bf 0 \\
Dust3r & 22.834 & 292.270 & 0.045 & 0.048 & 0.147 & 0.101 & 0.179 & 3 \\
Vistadream & 23.272 & \bf 226.189 & 0.037 & 0.049 & 0.131 & 0.092 & 0.160 & 4 \\
\our{} & \bf 23.559 & 234.560 & \bf 0.032 & \bf 0.012 & \bf 0.045 & \bf 0.032 & 0.097 & \bf 0 \\
\bottomrule
\end{tabular}
\vspace{-0.6cm}
\end{table}

\textbf{Generative Spatial Bridging} The objective of the first module is to establish sufficient semantic and geometric continuity between the two disjoint observations for reliable geometric initialization. To this end, we generate an intermediate ``anchor'' image between $I_1$ and $I_2$. We condition a Vision-Language Model (VLM), denoted as $\mathcal{P}_{vlm}$, on the visual inputs $(I_1, I_2)$ alongside a carefully engineered meta-prompt. The model $\mathcal{P}_{vlm}$ acts as a reasoning engine, analyzing the structural semantics of both views to output a highly descriptive, per-scene text prompt $\mathbf{p}$ that encapsulates the probable intermediate geometry. This text prompt, strictly alongside the source visual inputs $I_1$ and $I_2$, is then fed into a multi-conditional diffusion model, $\mathcal{P}_{diff}$. To mitigate the inherent stochasticity of diffusion processes, we sample $\mathcal{P}_{diff}$ to generate a diverse candidate pool of $M$ intermediate images, $\mathcal{C} = \{C^{1}, \dots, C^{M}\}$. Subsequently, a secondary VLM evaluator, $\mathcal{P}_{eval}$, is employed as a zero-shot reward model to select the optimal anchor image $I_{b} \in \mathcal{C}$ based on its spatial and semantic coherence with the source pair. 

Our framework leverages the robustness of modern geometry estimators to establish a coordinate system from a sparse triplet. By introducing the synthesized anchor, we formulate the expanded image set, 
$\mathcal{I}_{c} = \{I_1, I_{b}, I_{2}\} $. By synthesizing this anchor, we overcome the zero-overlap constraint, as the adjacent pairs $(I_1, I_{b})$ and $(I_{b}, I_2)$ now share non-empty visual intersections ($\Omega \neq \emptyset$), enabling valid geometric registration. However, the generated image may introduce hallucinations and geometric inconsistencies, making it unsuitable for direct reconstruction. To address these limitations, we employ the methods described in the subsequent sections.

\textbf{Robust Coarse 3D Reconstruction} While Generative Spatial Bridging establishes the necessary visual overlap, the generated intermediate frame inherently lacks strict epipolar consistency with the source views. Extracting a coherent 3D geometry from $\mathcal{I}_{c}$ thus requires a paradigm robust to generative noise. To address this we adopt an alternative paradigm based on unconstrained dense pointmap regression, denoted as $\mathcal{P}_{geo}$ \cite{wang2024dust3r}. By isolating the geometric lifting problem, $\mathcal{P}_{geo}$ treats pairwise reconstruction as a direct regression of dense 3D pointmaps rather than minimizing noise-sensitive 2D reprojection errors. We utilize $\mathcal{P}_{geo}$ to jointly optimize camera poses and scene geometry via a global alignment procedure directly in 3D coordinate space. This optimization acts as a spatial low-pass filter, effectively absorbing local generative contradictions. The resulting aligned point cloud is parameterized as an initial set of 3D Gaussians, forming our coarse geometric scaffold $\mathcal{S}_{c}$.

\textbf{Iterative Context Expansion and Consistency Optimization} Although the initial regression successfully maps our spatial bridge into a unified coordinate system, the scaffold $\mathcal{S}_{c}$ remains incomplete and contains floaters and visual inconsistencies. To transform this scaffold into a dense environment, we employ an iterative scene completion strategy inspired by the generative lifting techniques of VistaDream \cite{wang2025vistadream}.

To fill unobserved regions within the global coordinate space, we perform an iterative warp-and-inpaint procedure \cite{wang2025vistadream}. The system renders the partial scene from novel viewpoints along an interpolated trajectory. For viewpoints exhibiting significant holes, we utilize an RGB inpainting prior $\mathcal{P}_{inp}$, conditioned on the textual descriptors $\mathbf{p}$ obtained during the {\em Generative Spatial Bridging} stage, to synthesize missing visual context. To maintain structural integrity, a monocular metric depth estimator $\mathcal{P}_{dpt}$ is applied to the newly inpainted images. Crucially, this localized depth aligns with the existing depth constraints of $\mathcal{S}_{c}$ before new Gaussians are integrated into inpainted regions.

While sequential expansion completes the scene, it is prone to accumulating  drift \cite{wang2025vistadream}. To resolve these contradictions, we adopt a Multiview Consistency Sampling (MCS) approach \cite{wang2025vistadream}. Multiple rendered views from the updated scene are simultaneously injected with noise and processed through a Latent Consistency Model, $\mathcal{P}_{lcm}$ proposed in \cite{lcm}. By explicitly enforcing multi-view consistency constraints during the reverse diffusion sampling process, $\mathcal{P}_{lcm}$ rectifies diverging geometries and textures, yielding a topologically complete 3D representation \cite{wang2025vistadream}.

\begin{figure}[t]
  \centering
  \includegraphics[width=0.95\linewidth]{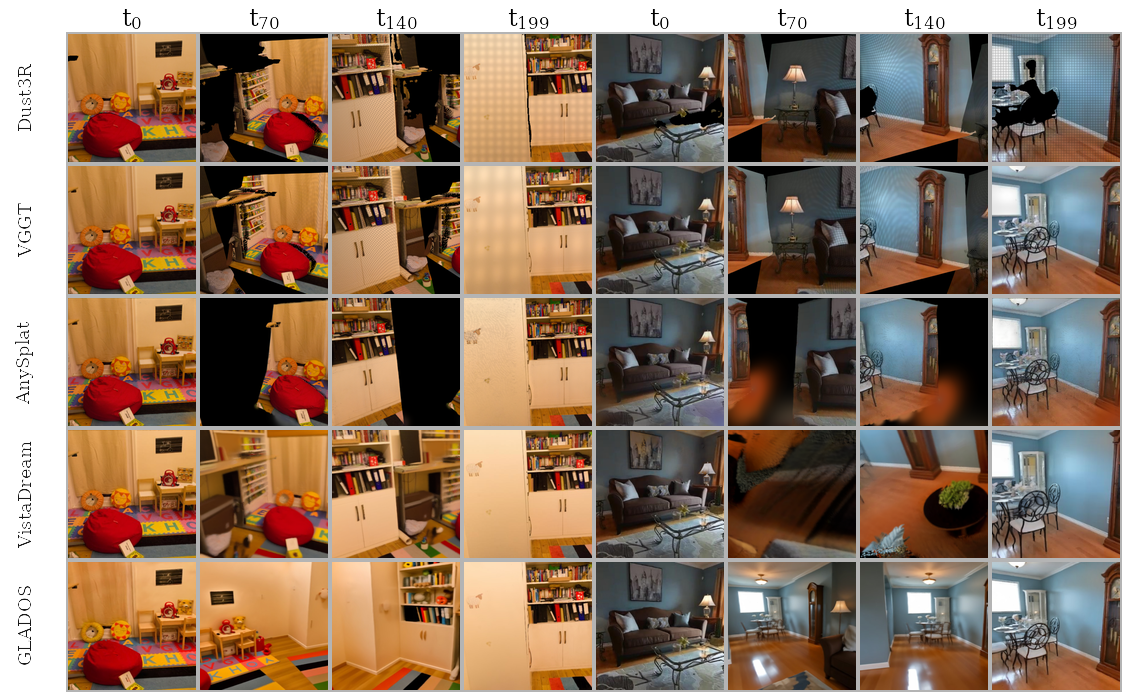}
  \caption{\textbf{Comparison of the reconstruction consistency between \our{} and other methods.} Purely reconstruction methods (\textit{Dust3R}, \textit{VGGT}, \textit{AnySplat}) fail to produce consistent scene, with major artifacts even for observed regions ($t_0$, $t_{199}$). \textit{VistaDream} produces less coherent results (left) or fails entirely in generating probable scene (right).}
  \label{fig:figure_5}
  \vspace{-0.7cm}
\end{figure}

\textbf{Generative 3D Refinement via Anchored Grid Inpainting} While the context expansion and consistency optimization successfully generate a fully enclosed 3D Gaussian field from our bridged views, the reliance on independent depth alignment and 2D diffusion often leaves residual geometric voids (holes) and localized textural blurring, Fig.\ref{fig:holes}. To resolve these final imperfections and elevate the reconstruction to strict metric accuracy, we introduce an iterative refinement loop that grounds generative inpainting directly to the original ground-truth observations.

To detect holes we reuse the camera trajectories generated during the {\em Iterative Context Expansion} phase. For a sampled viewpoint, we utilize the gsplat rasterizer \cite{gsplat} to render both the RGB image and the corresponding opacity ($\alpha$) map from the current scene state. The $\alpha$-channel explicitly identifies unobserved regions and residual holes (where $\alpha \approx 0$). To fill these gaps while maintaining consistency with the rest of the scene, we employ a grid-based inpainting strategy inspired by NeRFiller \cite{weber2024nerfiller}. We construct a $2 \times 2$ composite image grid containing two of our rendered novel views (which contain holes) and the two original disjoint ground-truth images ($I_1$ and $I_2$). The inpainting mask is constructed such that the ground-truth images remain fully preserved, while the novel views are masked only at pixels corresponding to missing $\alpha$-values. We process this composite grid through a diffusion prior $\mathcal{P}_{mgi}$. By forcing the network to process the unobserved novel views alongside the unmodified ground-truth anchors within the same receptive field, we enforce strict semantic and color consistency. To further stabilize the generation, we average the noise predictions across multiple denoising passes per step \cite{weber2024nerfiller}.

The output of $\mathcal{P}_{mgi}$ completes the visual context but may lack high-frequency fidelity due to the diffusion process. Therefore, we apply a super-resolution prior, $\mathcal{P}_{up}$, to enhance the inpainted regions. To lift high-resolution 2D patches back into 3D space, we pass the enhanced images through our metric depth estimator $\mathcal{P}_{dpt}$. Because monocular depth predictions can exhibit scale and shift ambiguities, we perform an affine alignment between the $\mathcal{P}_{dpt}$ prediction and the depth map rendered from the gaussian representation. Once aligned, new 3D Gaussian primitives are unprojected exclusively within the targeted masked regions and injected into the scene. Finally, the updated 3D Gaussian field undergoes a brief optimization phase to integrate the new primitives. This entire process: $\alpha$-rendering, anchored inpainting, super-resolution, depth alignment, and GS optimization is repeated iteratively until all residual holes and scene inconsistencies are eliminated, yielding the final, topologically complete reconstruction $\mathcal{S}_{fin}$.

%% file: sections/6-Benchmark.tex
\vspace{-0.2cm}
\section{Disjoint-View Benchmark}
\vspace{-0.2cm}

\begin{table}[t]
\caption{Quantitative comparison after augmenting all baseline methods with the proposed {\em Generative Spatial Bridging strategy.}}
\label{tab:average_metrics}
\centering
\small
\setlength{\tabcolsep}{5pt}
\begin{tabular}{lcccccccc}
\toprule
Method & CLIP $\uparrow$ & FID $\downarrow$ & MET3R $\downarrow$ & $\text{GeCo}_M \downarrow$ & $\text{GeCo}_D \downarrow$ & $\text{GeCo}_F \downarrow$ & Photo $\downarrow$ & Rec. Fail $\downarrow$ \\
\midrule
AnySplat & \bf 23.572 & 283.045 & 0.046 & 0.030 & 0.220 & 0.163 & 0.103 & \bf 0 \\
VGGT & 22.927 & 279.830 & 0.043 & 0.041 & 0.085 & 0.065 & 0.148 & \bf 0 \\
Dust3r & 22.969 & 267.033 & 0.048 & 0.057 & 0.215 & 0.141 & 0.173 & \bf 0 \\
Vistadream & 23.127 & \bf 206.169 & 0.034 & 0.048 & 0.107 & 0.085 & 0.158 & \bf 0 \\
\our{} & 23.559 & 234.560 & \bf 0.032 & \bf 0.012 & \bf 0.045 & \bf 0.032 & \bf 0.097 & \bf 0 \\
\bottomrule
\end{tabular}
\vspace{-0.3cm}
\end{table}

Existing reconstruction benchmarks are not suitable for evaluation in proposed setting due to overlapping input images. We therefore introduce a new disjoint-view benchmark tailored to this task.

\textbf{Dataset} We consider a range of factors, including indoor and outdoor scenes, diverse camera trajectories, and varying lighting conditions. We build our benchmark by sampling non-overlapping frame pairs from existing datasets such as Mip-NeRF 360 \cite{barron2022mipnerf360}, Tanks \& Temples \cite{tandt}, Deep Blending \cite{deepblending}, and RealEstate10K \cite{re10k}. Specifically, we include 2 scenes from Deep Blending, 2 scenes from Tanks \& Temples, 1 scene from MiP-NeRF 360, and 10 scenes from RealEstate10K. As these datasets do not cover all desired scenarios, we further augment the benchmark with 7 additional scenes collected specially for this work, resulting in a total of 22 scenes.

\textbf{Metrics} The proposed paradigm possesses sparse ground truth, thus metrics should be reference-free where possible. For assessing visual quality and coherence of novel views, we use FID \cite{fid} and CLIP Score \cite{hessel2022clipscorereferencefreeevaluationmetric}, with the latter using textual descriptions obtained during the first stage of the framework. To assess the geometric quality and coherence of the obtained reconstruction, we use the photometric error ($Photo$) between real image and reprojection of the generated view on the real camera. The error is calculated as pixel-wise MSE for pixels with valid depth, normalized by the number of valid pixels. Since, in our setting, references are sparse, we use GeCo (Geometric Consistency Metric) \cite{gu2025geco} and MEt3R (Measuring Multi-View Consistency in Generated Images) \cite{asim25met3r} as reference-free metrics. The former is designed to detect geometric deformation and occlusion-inconsistency artifacts in static scenes via \textit{motion map} ($GeCo_M$), \textit{structure map} ($GeCo_D$), and \textit{fused map} ($GeCo_F$). The latter compares feature maps from warped image pairs to compute a similarity score that is robust to view-dependent effects such as lighting changes. We additionally report the number of scenes for which each method failed to produce a valid 3D reconstruction (Rec. Fail). For details see appendix.

\textbf{Evaluation} Different reconstruction methods estimate camera poses using different internal parameterizations and coordinate systems, making direct comparison challenging. We generate a unified evaluation trajectory between the input views \(I_1\) and \(I_2\). Specifically, we interpolate between the corresponding camera poses using linear interpolation for translation and spherical linear interpolation (SLERP) for rotation, producing a trajectory of 200 intermediate camera poses for novel-views. This protocol enables fair comparison of visual quality, geometric consistency, and multi-view coherence independently of the camera estimation strategy employed by each method. We do not consider metrics for scenes where reconstruction failed.



%% file: sections/7-Results.tex
\vspace{-0.2cm}
\section{Experiments}
\vspace{-0.2cm}

We evaluate GLADOS on the proposed disjoint-view benchmark against representative state-of-the-art reconstruction and generative baselines, including AnySplat \cite{jiang2025anysplat}, VGGT \cite{wang2025vggt}, Dust3r \cite{wang2024dust3r}, and VistaDream \cite{wang2025vistadream}. The experiments are designed to assess both geometric consistency and generative plausibility under the zero-overlap setting introduced in this work. We first compare all methods in the challenging two-view reconstruction scenario, where only two spatially disjoint images are provided as input. Next, we evaluate the effect of integrating our {\em Generative Spatial Bridging} strategy into existing baselines. Finally, we perform ablation studies to analyze the contribution of each component of the proposed framework.

\textbf{Reconstruction from two views} We first evaluate all methods under the proposed disjoint-view setting, where each method receives only two non-overlapping input images. Quantitative results are reported in Table~\ref{tab:two_views} and qualitative in Fig. \ref{fig:figure_5}. Existing reconstruction approaches struggle in the absence of visual overlap. Methods relying primarily on geometric correspondence estimation, such as Dust3R and VGGT, exhibit reduced geometric consistency, while generative approaches such as VistaDream produce lower FID scores but frequently introduce structurally inconsistent scene completions. In contrast, \our{} achieves the best overall performance across most geometric metrics, including MEt3R, $\text{GeCo}_M$, $\text{GeCo}_D$, and $\text{GeCo}_F$, while also obtaining the highest CLIP score. Furthermore, \our{} does not exhibit reconstruction failures on any evaluated scene. 
Qualitative comparisons further demonstrate that purely generative approaches may hallucinate implausible structures and fail to maintain global scene consistency, as illustrated in Fig.~\ref{fig:figure_5}. 


\textbf{Reconstruction with Generative Spatial Bridging} We further evaluate the effect of the proposed Generative Spatial Bridging module by augmenting all baseline methods with the synthesized intermediate anchor image. Quantitative results are presented in Table~\ref{tab:average_metrics}. Introducing the additional generated view generally improves baseline methods' ability to establish coarse geometric alignment and reduces reconstruction failures. In particular, Dust3r no longer exhibits catastrophic reconstruction collapse when provided with the synthesized bridge image. However, the improvements remain inconsistent across methods and metrics. While some approaches benefit from the increased visual overlap, the generated anchor image may also introduce geometric ambiguities and hallucinated structures that propagate through the reconstruction process. 
In contrast, \our{} consistently achieves the strongest overall performance across both geometric and perceptual metrics, obtaining the best results in MEt3R, $\text{GeCo}_M$, $\text{GeCo}_D$, $\text{GeCo}_F$, and photometric error. 

\begin{table}[t]
\caption{\textbf{Ablation of \our{} components}. Bypassing spatial bridging (\textit{w/o Midpoint}) causes 4 total reconstruction failures. Omitting ICE\&CO severely degrades structural integrity, peaking photometric error. Halting before final refinement (\textit{w/o Refinement}) worsens GeCo and MEt3R metrics. The full method optimally balances robustness and geometric fidelity.}
\label{tab:ablation}
\centering
\small
\setlength{\tabcolsep}{4pt}
\begin{tabular}{lcccccccc}
\toprule
Method & CLIP $\uparrow$ & FID $\downarrow$ & MET3R $\downarrow$ & $\text{GeCo}_M \downarrow$ & $\text{GeCo}_D \downarrow$ & $\text{GeCo}_F \downarrow$ & Photo $\downarrow$ & Rec. Fail $\downarrow$ \\
\midrule
w/o Midpoint & 23.589 & 263.144 & 0.031 & 0.046 & 0.129 & 0.090 & 0.169 & 4 \\
w/o ICE\&CO & \bf 23.770 & 310.579 & 0.041 & 0.054 & 0.114 & 0.092 & 0.189 & \bf 0 \\
w/o Refinement & 23.127 & \bf 206.169 & 0.034 & 0.048 & 0.107 & 0.085 & 0.158 & \bf 0 \\
\our{} Full & 23.559 & 234.560 & \bf 0.032 & \bf 0.012 & \bf 0.045 & \bf 0.032 & \bf 0.097 & \bf 0 \\
\bottomrule
\end{tabular}
\vspace{-0.6cm}
\end{table}


\textbf{Ablation} To validate \our{}, we conduct a quantitative ablation study (Table \ref{tab:ablation}) and provide corresponding qualitative comparisons in Appendix C. Removing the {\em Generative Spatial Bridging} (\textit{w/o Midpoint}) prevents the dense pointmap regression from consistently finding valid correspondences thus causing reconstruction failures. Bypassing the iterative expansion and consistency optimization (\textit{w/o ICE\&CO}) forces the pipeline to apply the final refinement directly to the coarse Gaussian scaffold initialized from the sparse triplet. While this preserves the original input images better, reflected by the higher CLIP score, it causes less coherent scene geometry, visible across GeCo, MEt3R, and FID metrics. Alternatively, halting the pipeline before our anchored grid inpainting (\textit{w/o Refinement}) leaves residual geometric voids and misalignment. Although this unrefined output produces high-quality individual renders (yielding a strong FID score), it lacks the strict geometric consistency and metric accuracy of our complete approach, leading to higher MEt3R, GeCo, and photometric errors. Finally, the full \our{} framework integrates topological bridging, dense structural priors, and strict metric grounding to ensure robust scene initialization and achieve the best performance across all geometry-grounded metrics.

%% file: sections/8-Conclusions.tex
\vspace{-0.2cm}
\section{Conclusions}
\vspace{-0.2cm}

We introduced Generative Reconstruction from Disjoint Views, a new reconstruction paradigm that addresses the fundamentally challenging setting of zero-overlap image observations. Unlike traditional reconstruction pipelines, which rely on shared visual correspondences, the proposed task requires jointly satisfying strict geometric constraints in observed regions while plausibly synthesizing unseen intermediate space. To support this problem formulation, we presented a dedicated disjoint-view benchmark together with an evaluation protocol tailored to sparse-reference generative reconstruction. To address this challenge, we proposed GLADOS, a modular framework that combines foundation-model reasoning, generative spatial bridging, robust coarse geometric initialization, and iterative consistency-driven scene refinement. By integrating generative priors with explicit geometric regularization, GLADOS successfully reconstructs unified and geometrically coherent 3D scenes from spatially disjoint image pairs. Experimental results demonstrate that existing state-of-the-art reconstruction and generative methods struggle in the zero-overlap setting, frequently producing disconnected geometries or structurally inconsistent scene completions, whereas GLADOS achieves consistently stronger geometric coherence and reconstruction stability. \\
\textbf{Limitation} The proposed framework relies on generative priors, which may introduce semantically plausible but geometrically inaccurate structures in heavily unconstrained regions with insufficient observational constraints.

%% file: arxiva.bbl
\begin{thebibliography}{10}

\bibitem{mildenhall2020nerf}
Ben Mildenhall, Pratul~P. Srinivasan, Matthew Tancik, Jonathan~T. Barron, Ravi
  Ramamoorthi, and Ren Ng.
\newblock {NeRF: Representing Scenes as Neural Radiance Fields for View
  Synthesis}.
\newblock In {\em ECCV}, 2020.

\bibitem{kerbl20233d}
Bernhard Kerbl, Georgios Kopanas, Thomas Leimk{\"u}hler, and George Drettakis.
\newblock {3D Gaussian Splatting for Real-Time Radiance Field Rendering}.
\newblock {\em ACM Transactions on Graphics}, 42(4), 2023.

\bibitem{wang2024dust3r}
Shuzhe Wang, Vincent Leroy, Yohann Cabon, Boris Chidlovskii, and Jerome Revaud.
\newblock Dust3r: Geometric 3d vision made easy.
\newblock In {\em Proceedings of the IEEE/CVF conference on computer vision and
  pattern recognition}, pages 20697--20709, 2024.

\bibitem{jiang2025anysplat}
Lihan Jiang, Yucheng Mao, Linning Xu, Tao Lu, Kerui Ren, Yichen Jin, Xudong Xu,
  Mulin Yu, Jiangmiao Pang, Feng Zhao, et~al.
\newblock Anysplat: Feed-forward 3d gaussian splatting from unconstrained
  views.
\newblock {\em ACM Transactions on Graphics (TOG)}, 44(6):1--16, 2025.

\bibitem{smart2024splatt3r}
Brandon Smart, Chuanxia Zheng, Iro Laina, and Victor~Adrian Prisacariu.
\newblock Splatt3r: Zero-shot gaussian splatting from uncalibrated image pairs.
\newblock {\em arXiv preprint arXiv:2408.13912}, 2024.

\bibitem{cieslewski2018data}
Titus Cieslewski, Siddharth Choudhary, and Davide Scaramuzza.
\newblock Data-efficient decentralized visual slam.
\newblock In {\em 2018 IEEE international conference on robotics and automation
  (ICRA)}, pages 2466--2473. IEEE, 2018.

\bibitem{loy2010time}
Chen~Change Loy, Tao Xiang, and Shaogang Gong.
\newblock Time-delayed correlation analysis for multi-camera activity
  understanding.
\newblock {\em International Journal of Computer Vision}, 90(1):106--129, 2010.

\bibitem{heinly2015reconstructing}
Jared Heinly, Johannes~L Schonberger, Enrique Dunn, and Jan-Michael Frahm.
\newblock Reconstructing the world* in six days*(as captured by the yahoo 100
  million image dataset).
\newblock In {\em Proceedings of the IEEE conference on computer vision and
  pattern recognition}, pages 3287--3295, 2015.

\bibitem{lu2025align3r}
Jiahao Lu, Tianyu Huang, Peng Li, Zhiyang Dou, Cheng Lin, Zhiming Cui, Zhen
  Dong, Sai-Kit Yeung, Wenping Wang, and Yuan Liu.
\newblock Align3r: Aligned monocular depth estimation for dynamic videos.
\newblock In {\em Proceedings of the Computer Vision and Pattern Recognition
  Conference}, pages 22820--22830, 2025.

\bibitem{yang2025fast3r}
Jianing Yang, Alexander Sax, Kevin~J Liang, Mikael Henaff, Hao Tang, Ang Cao,
  Joyce Chai, Franziska Meier, and Matt Feiszli.
\newblock Fast3r: Towards 3d reconstruction of 1000+ images in one forward
  pass.
\newblock In {\em Proceedings of the Computer Vision and Pattern Recognition
  Conference}, pages 21924--21935, 2025.

\bibitem{chen2025easi3r}
Xingyu Chen, Yue Chen, Yuliang Xiu, Andreas Geiger, and Anpei Chen.
\newblock Easi3r: Estimating disentangled motion from dust3r without training.
\newblock In {\em Proceedings of the IEEE/CVF International Conference on
  Computer Vision}, pages 9158--9168, 2025.

\bibitem{wang2025vistadream}
Haiping Wang, Yuan Liu, Ziwei Liu, Wenping Wang, Zhen Dong, and Bisheng Yang.
\newblock Vistadream: Sampling multiview consistent images for single-view
  scene reconstruction.
\newblock In {\em Proceedings of the IEEE/CVF International Conference on
  Computer Vision}, pages 26772--26782, 2025.

\bibitem{chung2025luciddreamer}
Jaeyoung Chung, Suyoung Lee, Hyeongjin Nam, Jaerin Lee, and Kyoung~Mu Lee.
\newblock Luciddreamer: Domain-free generation of 3d gaussian splatting scenes.
\newblock {\em IEEE Transactions on Visualization \& Computer Graphics},
  (01):1--12, 2025.

\bibitem{zhao2025dreamscape}
Yueming Zhao, Xuening Yuan, Hongyu Yang, and Di~Huang.
\newblock Dreamscape: 3d scene creation via gaussian splatting joint
  correlation modeling.
\newblock In {\em 2025 IEEE International Conference on Multimedia and Expo
  (ICME)}, pages 1--6. IEEE, 2025.

\bibitem{yu2025viewcrafter}
Wangbo Yu, Jinbo Xing, Li~Yuan, Wenbo Hu, Xiaoyu Li, Zhipeng Huang, Xiangjun
  Gao, Tien-Tsin Wong, Ying Shan, and Yonghong Tian.
\newblock Viewcrafter: Taming video diffusion models for high-fidelity novel
  view synthesis.
\newblock {\em IEEE Transactions on Pattern Analysis and Machine Intelligence},
  2025.

\bibitem{zhang2025world}
Qihang Zhang, Shuangfei Zhai, Miguel Angel~Bautista Martin, Kevin Miao,
  Alexander Toshev, Joshua Susskind, and Jiatao Gu.
\newblock World-consistent video diffusion with explicit 3d modeling.
\newblock In {\em Proceedings of the Computer Vision and Pattern Recognition
  Conference}, pages 21685--21695, 2025.

\bibitem{erdelj2017wireless}
Milan Erdelj, Micha{\l} Kr{\'o}l, and Enrico Natalizio.
\newblock Wireless sensor networks and multi-uav systems for natural disaster
  management.
\newblock {\em Computer Networks}, 124:72--86, 2017.

\bibitem{recchiuto2018post}
Carmine~Tommaso Recchiuto and Antonio Sgorbissa.
\newblock Post-disaster assessment with unmanned aerial vehicles: A survey on
  practical implementations and research approaches.
\newblock {\em Journal of Field Robotics}, 35(4):459--490, 2018.

\bibitem{tian2022kimera}
Yulun Tian, Yun Chang, Fernando~Herrera Arias, Carlos Nieto-Granda, Jonathan~P
  How, and Luca Carlone.
\newblock Kimera-multi: Robust, distributed, dense metric-semantic slam for
  multi-robot systems.
\newblock {\em IEEE transactions on robotics}, 38(4), 2022.

\bibitem{lajoie2020door}
Pierre-Yves Lajoie, Benjamin Ramtoula, Yun Chang, Luca Carlone, and Giovanni
  Beltrame.
\newblock Door-slam: Distributed, online, and outlier resilient slam for
  robotic teams.
\newblock {\em IEEE Robotics and Automation Letters}, 5(2):1656--1663, 2020.

\bibitem{schmuck2019ccm}
Patrik Schmuck and Margarita Chli.
\newblock Ccm-slam: Robust and efficient centralized collaborative monocular
  simultaneous localization and mapping for robotic teams.
\newblock {\em Journal of Field Robotics}, 36(4):763--781, 2019.

\bibitem{evangelidis2021revisiting}
Georgios Evangelidis and Branislav Micusik.
\newblock Revisiting visual-inertial structure-from-motion for odometry and
  slam initialization.
\newblock {\em IEEE Robotics and Automation Letters}, 6(2):1415--1422, 2021.

\bibitem{chang2017matterport3d}
Angel Chang, Angela Dai, Thomas Funkhouser, Maciej Halber, Matthias Niessner,
  Manolis Savva, Shuran Song, Andy Zeng, and Yinda Zhang.
\newblock Matterport3d: Learning from rgb-d data in indoor environments.
\newblock {\em arXiv preprint arXiv:1709.06158}, 2017.

\bibitem{schonberger2016structure}
Johannes~L Schonberger and Jan-Michael Frahm.
\newblock Structure-from-motion revisited.
\newblock In {\em Proceedings of the IEEE conference on computer vision and
  pattern recognition}, pages 4104--4113, 2016.

\bibitem{yang2023freenerf}
Jiawei Yang, Marco Pavone, and Yue Wang.
\newblock Freenerf: Improving few-shot neural rendering with free frequency
  regularization.
\newblock In {\em Proceedings of the IEEE/CVF conference on computer vision and
  pattern recognition}, pages 8254--8263, 2023.

\bibitem{hollein2023text2room}
Lukas H{\"o}llein, Ang Cao, Andrew Owens, Justin Johnson, and Matthias
  Nie{\ss}ner.
\newblock Text2room: Extracting textured 3d meshes from 2d text-to-image
  models.
\newblock In {\em Proceedings of the IEEE/CVF International Conference on
  Computer Vision}, pages 7909--7920, 2023.

\bibitem{shi2023mvdream}
Yichun Shi, Peng Wang, Jianglong Ye, Mai Long, Kejie Li, and Xiao Yang.
\newblock Mvdream: Multi-view diffusion for 3d generation.
\newblock {\em arXiv preprint arXiv:2308.16512}, 2023.

\bibitem{agarwal2011building}
Sameer Agarwal, Yasutaka Furukawa, Noah Snavely, Ian Simon, Brian Curless,
  Steven~M Seitz, and Richard Szeliski.
\newblock Building rome in a day.
\newblock {\em Communications of the ACM}, 54(10):105--112, 2011.

\bibitem{martin2021nerf}
Ricardo Martin-Brualla, Noha Radwan, Mehdi~SM Sajjadi, Jonathan~T Barron,
  Alexey Dosovitskiy, and Daniel Duckworth.
\newblock Nerf in the wild: Neural radiance fields for unconstrained photo
  collections.
\newblock In {\em Proceedings of the IEEE/CVF conference on computer vision and
  pattern recognition}, pages 7210--7219, 2021.

\bibitem{zhang2025transplat}
Chuanrui Zhang, Yingshuang Zou, Zhuoling Li, Minmin Yi, and Haoqian Wang.
\newblock Transplat: Generalizable 3d gaussian splatting from sparse multi-view
  images with transformers.
\newblock In {\em Proceedings of the AAAI Conference on Artificial
  Intelligence}, volume~39, pages 9869--9877, 2025.

\bibitem{wang2025vggt}
Jianyuan Wang, Minghao Chen, Nikita Karaev, Andrea Vedaldi, Christian
  Rupprecht, and David Novotny.
\newblock Vggt: Visual geometry grounded transformer.
\newblock In {\em Proceedings of the Computer Vision and Pattern Recognition
  Conference}, pages 5294--5306, 2025.

\bibitem{kirillov2023segment}
Alexander Kirillov, Eric Mintun, Nikhila Ravi, Hanzi Mao, Chloe Rolland, Laura
  Gustafson, Tete Xiao, Spencer Whitehead, Alexander~C Berg, Wan-Yen Lo, et~al.
\newblock Segment anything.
\newblock In {\em Proceedings of the IEEE/CVF international conference on
  computer vision}, pages 4015--4026, 2023.

\bibitem{weber2024nerfiller}
Ethan Weber, Aleksander Holynski, Varun Jampani, Saurabh Saxena, Noah Snavely,
  Abhishek Kar, and Angjoo Kanazawa.
\newblock Nerfiller: Completing scenes via generative 3d inpainting.
\newblock In {\em Proceedings of the IEEE/CVF conference on computer vision and
  pattern recognition}, pages 20731--20741, 2024.

\bibitem{asim2026scenetok}
Mohammad Asim, Christopher Wewer, and Jan~Eric Lenssen.
\newblock Scenetok: A compressed, diffusable token space for 3d scenes.
\newblock {\em arXiv preprint arXiv:2602.18882}, 2026.

\bibitem{lcm}
Simian Luo, Yiqin Tan, Longbo Huang, Jian Li, and Hang Zhao.
\newblock Latent consistency models: Synthesizing high-resolution images with
  few-step inference, 2023.

\bibitem{gsplat}
Vickie Ye, Ruilong Li, Justin Kerr, Matias Turkulainen, Brent Yi, Zhuoyang Pan,
  Otto Seiskari, Jianbo Ye, Jeffrey Hu, Matthew Tancik, and Angjoo Kanazawa.
\newblock gsplat: An open-source library for gaussian splatting.
\newblock {\em Journal of Machine Learning Research}, 26(34):1--17, 2025.

\bibitem{barron2022mipnerf360}
Jonathan~T. Barron, Ben Mildenhall, Dor Verbin, Pratul~P. Srinivasan, and Peter
  Hedman.
\newblock Mip-nerf 360: Unbounded anti-aliased neural radiance fields.
\newblock {\em CVPR}, 2022.

\bibitem{tandt}
Arno Knapitsch, Jaesik Park, Qian-Yi Zhou, and Vladlen Koltun.
\newblock Tanks and temples: Benchmarking large-scale scene reconstruction.
\newblock {\em ACM Transactions on Graphics (ToG)}, 36(4):1--13, 2017.

\bibitem{deepblending}
Peter Hedman, Julien Philip, True Price, Jan-Michael Frahm, George Drettakis,
  and Gabriel Brostow.
\newblock Deep blending for free-viewpoint image-based rendering.
\newblock {\em ACM Transactions on Graphics (ToG)}, 37(6):1--15, 2018.

\bibitem{re10k}
Tinghui Zhou, Richard Tucker, John Flynn, Graham Fyffe, and Noah Snavely.
\newblock Stereo magnification: Learning view synthesis using multiplane
  images.
\newblock {\em ACM Trans. Graph. (Proc. SIGGRAPH)}, 37, 2018.

\bibitem{fid}
Christian Szegedy, Vincent Vanhoucke, Sergey Ioffe, Jonathon Shlens, and
  Zbigniew Wojna.
\newblock Rethinking the inception architecture for computer vision.
\newblock {\em CoRR}, abs/1512.00567, 2015.

\bibitem{hessel2022clipscorereferencefreeevaluationmetric}
Jack Hessel, Ari Holtzman, Maxwell Forbes, Ronan~Le Bras, and Yejin Choi.
\newblock Clipscore: A reference-free evaluation metric for image captioning,
  2022.

\bibitem{gu2025geco}
Leslie Gu, Junhwa Hur, Charles Herrmann, Fangneng Zhan, Todd Zickler, Deqing
  Sun, and Hanspeter Pfister.
\newblock Geco: A differentiable geometric consistency metric for video
  generation.
\newblock {\em arXiv preprint arXiv:2512.22274}, 2025.

\bibitem{asim25met3r}
Mohammad Asim, Christopher Wewer, Thomas Wimmer, Bernt Schiele, and Jan~Eric
  Lenssen.
\newblock Met3r: Measuring multi-view consistency in generated images.
\newblock In {\em IEEE/CVF Computer Vision and Pattern Recognition ({CVPR})},
  2025.

\bibitem{gemini}
{The Gemini Team}.
\newblock Gemini 3.1 pro: A smarter model for your most complex tasks, 2026.
\newblock Accessed: 2026-05-06.

\bibitem{nanobanana}
Naina Raisinghani.
\newblock Nano banana 2: Combining pro capabilities with lightning-fast speed,
  2026.
\newblock Google DeepMind blog, accessed 2026-05-06.

\bibitem{qwen}
Peng Wang, Shuai Bai, Sinan Tan, Shijie Wang, Zhihao Fan, Jinze Bai, Keqin
  Chen, Xuejing Liu, Jialin Wang, Wenbin Ge, Yang Fan, Kai Dang, Mengfei Du,
  Xuancheng Ren, Rui Men, Dayiheng Liu, Chang Zhou, Jingren Zhou, and Junyang
  Lin.
\newblock Qwen2-vl: Enhancing vision-language model's perception of the world
  at any resolution.
\newblock {\em arXiv preprint arXiv:2409.12191}, 2024.

\bibitem{fooocus}
Lvmin Zhang and Fooocus contributors.
\newblock Fooocus: Focus on prompting and generating.
\newblock \url{https://github.com/lllyasviel/Fooocus}, 2024.
\newblock Accessed: 2026-05-06.

\bibitem{depthpro}
Aleksei Bochkovskii, Ama\"{e}l Delaunoy, Hugo Germain, Marcel Santos, Yichao
  Zhou, Stephan~R. Richter, and Vladlen Koltun.
\newblock Depth pro: Sharp monocular metric depth in less than a second.
\newblock In {\em International Conference on Learning Representations}, 2025.

\bibitem{sd}
Robin Rombach, Andreas Blattmann, Dominik Lorenz, Patrick Esser, and Björn
  Ommer.
\newblock High-resolution image synthesis with latent diffusion models, 2021.

\bibitem{realesrgan}
Xintao Wang, Liangbin Xie, Chao Dong, and Ying Shan.
\newblock Real-esrgan: Training real-world blind super-resolution with pure
  synthetic data.
\newblock In {\em International Conference on Computer Vision Workshops
  (ICCVW)}.

\end{thebibliography}
